\documentclass[letterpaper]{article} 
\usepackage{aaai2026}  
\usepackage{times}  
\usepackage{helvet}  
\usepackage{courier}  
\usepackage[hyphens]{url}  
\usepackage{graphicx} 
\usepackage{natbib}  
\usepackage{amsmath}
\usepackage{caption} 
\usepackage{algorithm}
\usepackage{algorithmic}
\usepackage{booktabs}

\usepackage{newfloat}
\usepackage{listings}
\DeclareCaptionStyle{ruled}{labelfont=normalfont,labelsep=colon,strut=off} 
\floatstyle{ruled}
\newfloat{listing}{tb}{lst}{}
\floatname{listing}{Listing}
\title{Toward a Theory of Value in AI Alignment}
\author{
    Andrew Smart \textsuperscript{\rm 1}, 
    Shazeda Ahmed \textsuperscript{\rm 2},
    Jackie Kay \textsuperscript{\rm 3},
    Jimmy Tobin \textsuperscript{\rm 1},
    Kris Shrishak \textsuperscript{\rm 4},
    Abeba Birhane \textsuperscript{\rm 5}
}
\affiliations{
    \textsuperscript{\rm 1}Google Research\\
    \textsuperscript{\rm 2}UCLA\\
    \textsuperscript{\rm 3}Google Deep Mind\\
    \textsuperscript{\rm 1}Google Research\\
    \textsuperscript{\rm 4}OECD\\
    \textsuperscript{\rm 5}Trinity College Dublin\\
    andrewsmart@google.com

}

\usepackage{bibentry}

\begin{document}

\maketitle

\begin{abstract}
Can AI systems be aligned to human values? The popularization of large language models (LLMs) and multi-modal foundation models has seen a rise harms spanning from toxic speech and hallucinations to AI agents executing unauthorized actions. Within the field of AI safety, these harmful instances are often framed as “the alignment problem,” of models being “misaligned” with human values. Researchers have responded by pursuing applied and theoretical AI “value alignment” efforts, often without specifying what they mean by human values. How does the field of AI value alignment conceive of human values? How are these conceptions of values technically operationalized and evaluated? What does the emergent theory of value from this field signify for the future of AI?

We annotated 94 AI value alignment research papers to discern their implicit theory of values in AI. The majority do not define values, relying heavily on “preferences” as a stand-in that runs the risk of reducing complex, culturally situated concepts down to binary choices. As researchers dispense with using human annotators for model training and evaluation, turning instead to synthetic data and LLM-as-a-judge approaches to aligning and evaluating models, we identify the potential to close off alternative methods for contesting and enacting values in foundation models. In making AI value alignment’s philosophical commitments explicit, we seek to bring greater specificity and under-explored perspectives into the debate on whether and how AI can address human values.
\end{abstract}


\section{Introduction}

Large language models (LLMs) and multi-modal AI models have transformed almost all domains of social, economic, and scientific practices from research projects to consumer products 
~\cite{gillespie2024ai}. One area of concern is that, as these systems become more complex and widespread, it can be more difficult to understand, predict, and control their actions, potentially leading them to exhibit harmful outputs that were unintended by the models' human creators. However, as Guest et al point out, model designers do know the mechanistic structure of these models because they designed and built them, and to claim ignorance of how neural networks function is a form of mysticism \cite{guest2026does}. Within the research field of AI safety, a wide swath of instances of models deviating from human intentions to cause harm is referred to as "the alignment problem," often treated as the most important unsolved problem in machine learning. In recent years, AI alignment research has emerged as a leading strand of AI safety research designed to enact a contested belief: that catastrophes where misaligned AI systems will evade human control and inflict irreversible damage can be averted if ML systems are \lq aligned\rq with human values ~\cite{gabriel_artificial_2020,ji2023ai,askell2021general, russell_human_2019}. A large technical literature now concerns "value alignment" in AI, yet fundamental questions about what human values are, and with whose values AI should be aligned, remain under-examined. In this paper, we conducted a analysis of 94 AI alignment papers to reveal the emerging literature's underlying theory of what constitutes human values in ML models. To do so, we developed a rubric that draws from the study of values in anthropology, philosophy and other social sciences, which we then used to assess the papers in our dataset. We build on recent recognition of these foundational problems in AI alignment research~\cite{zhi-xuan_beyond_2024, arzberger2024nothing}. Before norms and practices in the value alignment subfield begin to ossify, our paper serves as a critical intervention to reassess the epistemic claims, methods, and oversights that are coming to define this work, and as an invitation to include the perspectives of more interdisciplinary scholarship and perspectives. 

\section{Related Work}
\subsection{What is value alignment?}
Why should alignment be necessary in the first place? Fundamentally, the training objective of LLMs is to correctly predict how an incomplete piece of text will continue. However, success in this training objective does not mean that the model can perform well in any arbitrary downstream task, or that the model is \lq safe\rq to use~\cite{hooker2025slow}. 
Value alignment focuses on \lq steering\rq an AI model toward outputs that humans (and increasingly other AI models~\cite{sharma2024critical}) judge as in accordance with certain preferences, e.g., helpfulness, \lq not being racist\rq, safety, or fairness, through fine-tuning methods including reinforcement learning from human feedback (RLHF)~\cite{christiano_deep_2017,bai_training_2022,chaudhari2024rlhf}, inverse reinforcement learning (IRL)~\cite{hadfield-menell_cooperative_2016}, direct preference optimization (DPO)~\cite{amirloo2024understanding}, Bradley-Terry-based models, and related techniques ~\cite{bai_training_2022, hofmann_ai_2024, hadfield-menell_cooperative_2016, ouyang2022training, sun2024rethinking, chaudhari2024rlhf}. Within this paradigm, value alignment is reformulated as encoding the preferences demonstrated by a human into a reward function, and updating the parameters of the language model to produce output that maximizes this reward. The assumption is that the AI can learn what human values and utility functions are by observing their behavior \cite{hadfield-menell_cooperative_2016}. 

 In public-facing discourse, AI safety researchers project the future possibility that misaligned foundation models could invoke damage ranging from unauthorized financial transactions to taking over power grids. In worst-case scenarios referred to as \lq existential risk\rq (x-risk), some in the AI safety community fear that rogue LLMs could kill most of the human population ~\cite{kasirzadeh2024two}. Others believe that such an event could lead to human extinction~\cite{nauer2025existential}. On the flip side, many of these same people claim that if AI systems can be \lq aligned\rq, this could bring about a utopia where human beings would no longer have to work. Yet despite the high stakes that alignment researchers place on \lq solving\rq alignment, research on value alignment rarely makes an effort to define values, let alone to center the analysis of pre-existing human values as part of AI safety research. 


Regardless of whether one believes in x-risk scenarios, value alignment is not merely a theoretical issue to be resolved within the machine learning community. The current state of value alignment research has real-world implications. Corporations are marketing their \lq value-aligned\rq AI products to governments around the world. For example, the European Parliament uses Anthropic's LLMs to provide access to its archives. The Parliament's choice was based on Anthropic's claim that its Constitutional AI is value-aligned. Although there has been no independent evaluation of Anthropic's approach, the claim of value alignment has been sufficient for governments to accept that these LLMs are legally compliant~\cite{shrishak_2025}. Elsewhere, governments are funding alignment research, such as the GBP £15 million Alignment Project within the UK's AI Security Institute~\cite{UKAISI2025}.

To date, value alignment is predominantly a technical field even though its underlying premise is sociotechnical. Yet foundation models function as epistemic technology that embeds normative models of knowledge within computational design \cite{alvarado2023ai} while also serving as a product of the social and cultural milieus that produce them. Alignment proponents advocate for encoding human values, moral principles or objectives in AI systems, most often via incorporating human feedback. For instance, Anthropic, the AI firm that produced the chatbot Claude, has published influential research positing that LLMs should exhibit traits such as helpfulness, honesty and harmlessness (HHH)~\cite{askell2021general}. Others in the field attempt to guide LLMs to adhere to principles such as robustness, interpretability, controllability, and ethicality~\cite{ji2023ai}, or safety, quality and groundedness~\cite{thoppilan2022lamda}. Many researchers default to HHH and other off-the-shelf values that are associated with training datasets and benchmarks, rarely questioning whether these values are indeed enacted in these technical artifacts.


\subsection{Alignment and its discontents}
As alignment crystallizes into a core research wing of AI safety, it is crucial to examine the roots of this work through fundamental questions around the very idea of value alignment. What are the core assumptions and feasibility of this project? Is it a meaningful avenue for fair, transparent, and accountable AI development? Most alignment research has examined how to encode moral values into AI models to guide their behavior~\cite{bergman2024stela}.  The question of \textit{what moral values are}, however, is left undefined or under-specified~\cite{gabriel_artificial_2020}. Existing alignment approaches tend to rely on universal framings of human values that obscure the question of which values the systems should capture and align with, despite the variety of operational situations in which these systems are deployed ~\cite{arzberger2024nothing}.  

Value alignment efforts suffer from vagueness, lack of internal consistency, and lack of commitment to establishing clear guidelines for how to determine what is acceptable AI system behavior ~\cite{lindstrom2024ai,arvan2024interpretability}. The term \lq values\rq is often left undefined in alignment literature, its meaning assumed to stem from taken-for-granted background knowledge. Moreover, crucial questions around \lq whose values\rq are hardly addressed. The absence of transparency around these implicit \lq values\rq, how they are derived, and how they are implemented in and itself a problem. Even though the importance of the diversity of views (\lq pluralism\rq) is sometimes acknowledged, "to date the research and policy proposals coming out of this [AI safety] community have converged around a narrow set of technical solutions that do not engage with work that falls outside of the [its] ideological and disciplinary boundaries"~\cite{ahmed2024field}.

Tech industry alignment research markets the mission of alignment as "benefiting humanity" without acknowledging the vast diversity of the human experience. Major AI firms have in-house alignment experts, blogs ~\cite{openai2025blog}, and teams ~\cite{anthropic2025blog}. In response to how corporate approaches to AI alignment reverberate across the field, one critique of alignment's shortcomings pertains to the profit motive driving the companies behind the most widely used models: "the existence of financial incentives means that alignment work often turns into product development in disguise rather than actually making progress on mitigating long-term harms"~\cite{dai2025artificiality}.

Furthermore, ~\cite{lindstrom2024ai} contend that even when objectives such \lq harmlessness\rq are identified as a chief aim, the nuanced nature of harm is oversimplified and operationalized in a way that is internally inconsistent and vague. Due to superficial understanding of ethical behavior in AI systems, what is often sought is the "least harmful option rather than striving to understand the deeper roots of harm and addressing these to prevent it" ~\cite{lindstrom2024ai}. The phrase 'value alignment' lacks an agreed-upon meaning. Measurements of the degree to which a system is value-aligned are subjective at best~\cite{khlaaf2023toward}. According to~\citet{kirk2023empty}, 'alignment' is an empty signifier that serves as a "rhetorical placeholder for an aspirational conceptualisation". A lack of cohesion according to~\citet{khlaaf2023toward} has led to contradictory approaches that conflate safety properties with system requirements. 

The narrow focus on internal technical components and emphasis on the mathematical formulation, including the objective or reward function is another limitation of value alignment efforts. Contrary to common assumptions within value alignment research, harms and failures due to accidents are not \textit{unanticipated emergent behaviors} but rather a \textit{byproduct of basic design choice}, resource requirements data, and compute requirements, as well as designers' and developers' API model delivery decisions~\cite{raji2023concrete,dobbe2022system}. Given that AI systems are sociotechnical systems that consist of technical AI artifacts, human agents, and institutions, comprehensively addressing safety issues requires thorough measures that include addressing how a system is used in practice and interacts with other (human) agents, systems, and its broader environment~\cite{dobbe2022system}. Value alignment's current technical focus, along with the framing that treats 'misaligned' AI as a danger to human survival if not 'controlled and steered,' serves those who are developing and deploying AI systems to evade accountability~\cite{gebru2024tescreal}. 

\subsection{Preferences are not all you need}
Although researchers are beginning to contend with problematic assumptions and conceptions of value underlying AI value alignment~\cite{johnson_are_2023, lindstrom2024ai, casper2023open}, one overlooked flaw we explicate is the field's implicit grounding in economic theory. In practice, value alignment's reliance on economic theory manifests as an emphasis on rational choice theory and "expected utility" (which is the average "reward" or "payoff" of a decision or choice)~\cite{becker1976economic, russell1995artificial, chibnik_anthropology_2011}. Model developers' decision to substitute values with preferences in reinforcement learning from human feedback (RLHF), direct preference optimization (DPO)~\cite{rafailov2023direct, amirloo2024understanding} or via AI feedback (RLAIF)~\cite{ji2023ai} is likewise based on revealed preference theory from economics. This theory posits that an agent's values or goals can be inferred from observing their behavior--an assumption that has AI researchers have largely adopted without considering alternatives~\cite{casper2023open}. 
More recent approaches such as DPO obviate the need for a reward model, but nonetheless use preference datasets~\cite{rafailov2023direct}.

If the assumptions underlying economic theory are correct, this process should "align" AI with human values. However, relying on highly abstract, reductionist, fictionalized, and idealized concepts of rational agents in lieu of a deep understanding of human values has been central to AI for decades~\cite{russell1995artificial}. This dominant technical paradigm in AI alignment is what \citet{zhi-xuan_beyond_2024} term the "preferentist" approach to value alignment, where people's preferences are translated into data points to provide aggregate guidance about preferred outcomes~\cite{zhi-xuan_beyond_2024,gabriel_artificial_2020}. This approach has three core assumptions: 1) that preferences are an adequate representation of human values, 2) that rationality consists of maximizing the satisfaction of preferences, and 3) that aligning AI models to data about these preferences is a sound technique for making AI safe~\cite{zhi-xuan_beyond_2024}. 

A widely recognized problem is that within the preferentist paradigm, even with pluralist frameworks, current methods for fine-tuning language models from human preferences treat these preferences (and the unexamined values shaping the preferences) as if they were homogeneous and static~\cite{bakker2022fine, klassen_pluralistic_2024, sorensen_roadmap_2024, lindstrom2024ai}. 


More fundamentally, the very project of reliably aligning LLM behavior with human values has been criticized as provably impossible~\cite{arvan2024interpretability}. \citet{arvan2024interpretability} argues that not only is there persistent human disagreement over moral values and principles, both among the public and moral theorists, but also that LLMs are complex given that they contain finite series of observational data and an infinite number of ‘misaligned’ functions. In other words, there is always a vastly higher empirical probability that an LLM will diverge from what appears to be an ‘aligned function’ to some ‘misaligned’ one later. "‘Alignment’, thus is not a "solvable" engineering or safety problem, but rather a quixotic and potentially dangerous fantasy based on a series of philosophical misunderstandings about empirical evidence"~\cite{arvan2024interpretability}. 


\section{Statement of Contributions}

Motivated by the issues outlined in the above discussion, we construct and analyze a sample of AI value alignment literature to critically assess how researchers typically operationalize value. The question we raise in this paper is: what implicit theories of human values underpin the field of AI value alignment? 

Our contributions are: 
\begin{itemize}
    \item We develop a framework for analyzing the theory of value in AI alignment, inspired by interdisciplinary literature in the social sciences and philosophy.
    \item Using this framework, we conduct a qualitative study of how value is theorized from a snowballed sample of 100 commonly cited AI alignment papers, seeded by four canonical papers in the field.
    \item We form an initial hypothesis of where AI value alignment’s theory of value sits, based on our critical observations of the field.
    \item We briefly outline alternative theories of human values as potential paths forward for thinking about how to mitigate harms from large AI systems.
\end{itemize}


\section{Methods}
\label{Methods}

 To characterize the dominant paradigm of value in AI research, we conducted a study of influential AI alignment papers, quantifying their philosophical positions using a scoring rubric developed from our above provocations.

We chose 4 canonical ‘seed papers’ on AI value alignment based on our expert judgment and high Google Scholar citations counts ($>1,000$ ). The seed papers have been highly influential in defining the field of technical AI alignment. Regarding the sample size and representativeness, we note that in the social sciences, when conducting close readings and qualitative analyses of texts, 94 papers is an acceptable sample. We did not use automated methods to evaluate these papers.

These 4 seed papers are:
\begin{itemize}
    \item \textbf{Deep Reinforcement Learning from Human Preferences} \cite{christiano2017deep}, 
    \item \textbf{A General Language Assistant as a Laboratory for Alignment} \cite{askell2021general}, 
    \item \textbf{Training a Helpful and Harmless Assistant with Reinforcement Learning from Human Feedback} \cite{bai_training_2022}, 
    \item \textbf{Training language models to follow instructions with human feedback} \cite{ouyang2022training}
\end{itemize}

\subsection{Snowball sampling}
Based on the 4 seed papers, we used a snowball sampling method \cite{parker2019snowball} to automatically download papers that met the inclusion criteria based on "value alignment". Using the Semantic Scholar API, all papers from 2016-24 that cite these 4, with an initial corpus of 576 papers.We sorted these from highest to lowest Google Scholar citations. We sorted the papers by Google Scholar citation count as a weak proxy for the quality and impact of each paper. During annotation, papers were discarded if they used the term 'value alignment' only in a cursory sense, rather than addressing values as a central consideration of the paper.After reaching consensus on which papers were about AI value alignment, out of the original 576 papers, we kept 94 for annotation. The sample size was also chosen due to resource constraints: querying Google Scholar citation counts is time-consuming, as is the paper annotation process itself. 


 Paper selection, scoring, and annotation was conducted by the authors. We used the annotation rubric to guide reading of the papers for scoring each question. We kept detailed qualitative notes in one column of the dataset, using comments and selecting illustrative quotes and passages from each paper to support our qualitative analysis of this literature. 

\subsection{Annotation rubric}
\emph{Rubric.} In order to characterize the theory of value underlying top cited AI alignment literature, we developed a rubric of questions based on a review of relevant literature in philosophy, sociology and anthropology that has not been considered within value alignment research. 
All questions are ternary: the answer can be one of two specific options, or "not applicable" if the answer is ambiguous or cannot be determined from the available information. 

Development of the rubric was iterative, responding to emergent themes we noticed early in the annotation process. For example, we began developing our rubric with the assumption that papers ostensibly about "value alignment" would engage with human values, and our initial rubric draft did not include the question of whether the papers under consideration even defined or engaged with human values. Yet from a first pass at annotating papers, it became clear that we needed to first include the question of whether the paper defined human values because so few papers met this expectation. We also realized that many of the papers in our sample relied on "autoraters" or LLMs as stand-ins for humans, and therefore we revised the rubric to include a question about whether actual humans were consulted about their values. 

\subsubsection{Justification for rubric questions}
One of the most fundamental distinctions philosophers make about values is between \textbf{monism} and \textbf{pluralism}. Monism is the view that there is a \textit{single value} that explains the value of everything (e.g.,
happiness, or pleasure) and that this single value can be ordered on a \textit{cardinal scale} or a hierarchy of ends \cite{korsgaard1986aristotle}. Strong monists claim that on this cardinal scale there are
units of pleasure (hedons or utils) and the
intervals between them is constant. Monism about value corresponds with utilitarianism, economic utility and rational choice theory \cite{mill2016utilitarianism, becker1976economic}. 

\textbf{Pluralism} in contrast holds that there is not just a single value, that human actions might have multiple final ends, and that there are multiple things that are good in and of themselves such as love, knowledge, piety, and justice \cite{haslanger_situated_2023, haslanger2022failures}. Pluralism recognizes that values are culturally relative and fundamentally vary across societies \cite{schmer2026culture}. Thus, a significant development in alignment research recently has been what we call "the pluralist turn," which seeks to develop approaches to accommodate diverse viewpoints and values \cite{sorensen_roadmap_2024, klassen_pluralistic_2024, kasirzadeh_plurality_2024}. 

We incorporated questions about the quantifiability and measurability of values, and assessed papers for whether they view values as measurable. The assumption that values are the equivalent of maximizing a utility function allows for the ostensible quantification of human values. From anthropology we adapted the evaluative distinction between \textbf{thin} and \textbf{thick} conceptions of values. Thick evaluative concepts and descriptions of human values view them not as mere abstractions, but as involving intentional, purposive detail that helps us understand those activities in their cultural and social contexts \cite{geertz_thick_2008}. 

Given the central role that economic theory plays in AI alignment, we also annotated papers for whether the alignment framework was rooted in utility theory.

\subsubsection{Rubric questions used to annotate the papers}
\begin{enumerate}
    \item Does the paper define or describe \textbf{value}? 
    \item Does the paper's theory of value subscribe to value \textbf{monism} or \textbf{pluralism}? Monism holds that there is a single, intrinsic dimension to value, while pluralism states that there is more than one dimension or consideration of value, which cannot be flattened.
    \item Does the paper consider values \textbf{measurable} or \textbf{immeasurable}? Measurable values can be quantified, operationalized, and recorded as data with negligible loss of precision, while immeasurable values cannot be precisely operationalized and quantified.
    \item Does the paper consider values as revealed \textbf{preferences}, or prescribed \textbf{principles}? Preferences are measured from behavioral decisions, and may involve some process of ranking preferred actions, outcomes, or things, while principles-based values correspond to abstract ideals or standards, e.g. valuing beauty, truth, knowledge, human rights declarations, etc.
    \item Is the approach to values \textbf{abstract} or \textbf{concrete}? Abstract approaches can be theoretical, idealized, philosophical, name no specific values in the paper but present a set of principles for extracting them, whereas concrete approaches may pertain to a specific existing system, culture, or group.
    \item Are values apparent in \textbf{individual} interactions, or formed from a \textbf{collective} of agents? That is, are values observed, measured, or determined from individual or one-on-one interactions, or do they arise over communities, cultures, nations, or groups?
    \item Are values \textbf{dynamic} or \textbf{static}? Dynamic values can change over time or in different situations, while static values do not change (or the change of values is not explained).
    \item Is the characterization of values \textbf{thin} or \textbf{thick}?
    We take a "thin" characterization of value to be entirely instrumental, evaluative and calculative, while a "thick" characterization is descriptive and situated in a societal context.
    \item Does the paper assume or explicitly state that values can emerge in AI autonomously from its creators?
    \item \textbf{Does the paper take the position that AI should follow human values}? A normative question which asks if it is morally justified, ethically correct, and/or strategically important for the field to develop AI systems which "uphold" values, resembling the processes by which humans uphold values (according to the paper's implicit or explicit theory of value).
    \item Does the paper identify \textbf{which humans'} values the research is aligning to?
    \item When preferences are elicited from human annotators, does the paper provide information about the \textbf{sample size of raters}?
    \item Does the paper use \textbf{utility maximization} as an approach to value alignment?
\end{enumerate}

\subsubsection{Annotation process}
We ensured that each paper was annotated by two annotators among the five of the co-authors of this paper. We used a Google form with the rubrics, and filled them in after reading each paper. We divided the papers into groups so that a subset of annotators read approximately fifty papers each. After an initial pass at reading the collected papers, we isolated disagreements and discussed them through resolutions and consensus. This was, however, not done for every disagreement and some disagreement remains among the annotations. However, we achieved a high degree of inter-annotator agreement, which averages overall around 85\%. 

\subsection{Qualitative and interpretive analysis}
In addition to scoring the papers according to our rubric for a quantitative estimate of the prevalence of the concepts, we also collected quotes from each paper if they addressed human values in non-mathematical terms. We did this in order to characterize the beliefs about human values stated, where applicable, by the authors of the papers we analyze. 

\section{Findings}
\label{Findings}
\subsection{Quantitative results}
\subsection{Inter-rater agreement}
In Table \ref{tab:rater-stats} we report measures of inter-rater agreement.

\begin{table*}[ht]
    \centering
    \caption{Inter-Rater Reliability Statistics}
    \label{tab:rater-stats}
    \begin{tabular}{l c c c c c}
        \toprule
        \textbf{Question} & \textbf{Mean $\kappa$} & \textbf{Mean PABAK} & \textbf{$\alpha$ (Alpha)} & \textbf{Agreement} & \textbf{\# Pairs} \\
        \midrule
        Are values measurable? & 0.541 & 0.775 & 0.269 & 88.8\% & 4 \\
        Revealed preferences or prescribed principles? & 0.467 & 0.710 & 0.373 & 85.5\% & 4 \\
        Sources for determining values & 0.411 & 0.167 & 0.023 & 58.3\% & 4 \\
        Abstract or Concrete approach & 0.378 & 0.619 & 0.235 & 81.0\% & 4 \\
        Uses utility maximization? & 0.365 & 0.685 & 0.428 & 84.2\% & 4 \\
        Can values emerge in AI? & 0.350 & 0.386 & 0.331 & 69.3\% & 4 \\
        States which humans' values? & 0.302 & 0.601 & 0.219 & 80.1\% & 4 \\
        Which humans' values? & 0.299 & 0.736 & 0.000 & 86.8\% & 4 \\
        Think vs Thick. & 0.239 & 0.056 & 0.073 & 52.8\% & 4 \\
        Does the paper define value? & 0.221 & 0.501 & 0.247 & 75.1\% & 4 \\
        Monism vs Pluralism & 0.218 & 0.079 & 0.349 & 54.0\% & 4 \\
        Values Static or Dynamic? & 0.133 & 0.021 & 0.204 & 51.0\% & 4 \\
        Should AI follow human values? & 0.090 & 0.138 & -0.015 & 56.9\% & 4 \\
        Size of rater pool & 0.056 & 0.275 & 0.348 & 63.7\% & 4 \\
        Individual vs Collective interactions & 0.028 & 0.215 & 0.286 & 60.8\% & 4 \\
        \bottomrule
    \end{tabular}
    \\[5pt] 
    \footnotesize 
    \textbf{Notes:} PABAK = $2 \times \text{Agreement} - 1$ (useful when class distributions are skewed); Krippendorff's $\alpha$ considers all raters simultaneously and handles missing data. Limitations of quantitative analysis: we report raw agreement, Cohen's kappa, Krippendorff's $\alpha$ and Bias-Adjusted Kappa (PABAK). Given that, we do expect high agreement on the binary yes/no questions, but some disagreement on extent or categorical questions; these results could indicate issues with design or an attribute of the setting in which we are not the research subjects. Some subjectivity may not be fully specified by the rubric questions. We may have different criteria for what counts as answers to our questions treating papers as the annotation item. We are not claiming that we have developed an objective survey \cite{aroyo2015truth}. We as the annotators are not the research subjects, which is one methodological assumption baked into inter-rater reliability; another is that the raters are interchangeable, which we are not.
\end{table*} 

\begin{figure*}[t]
    \centering
    \includegraphics[width=0.8\textwidth]{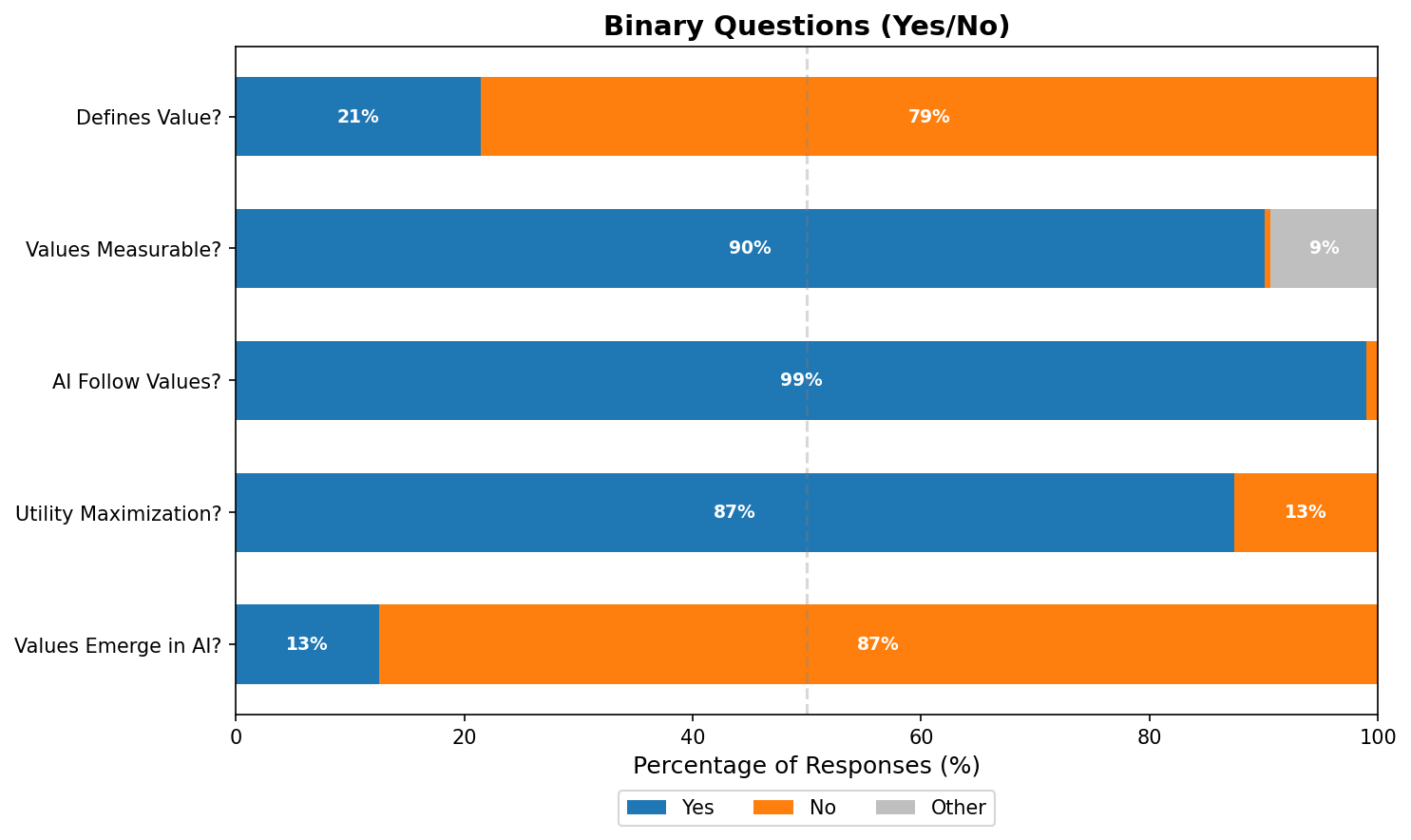}
    \caption{Summary of answers to our rubric aggregated across all raters for binary yes/no answers.}
    \label{fig:binary_answers}
\end{figure*}

\begin{figure*}[t]
    \centering
    \includegraphics[width=0.8\textwidth]{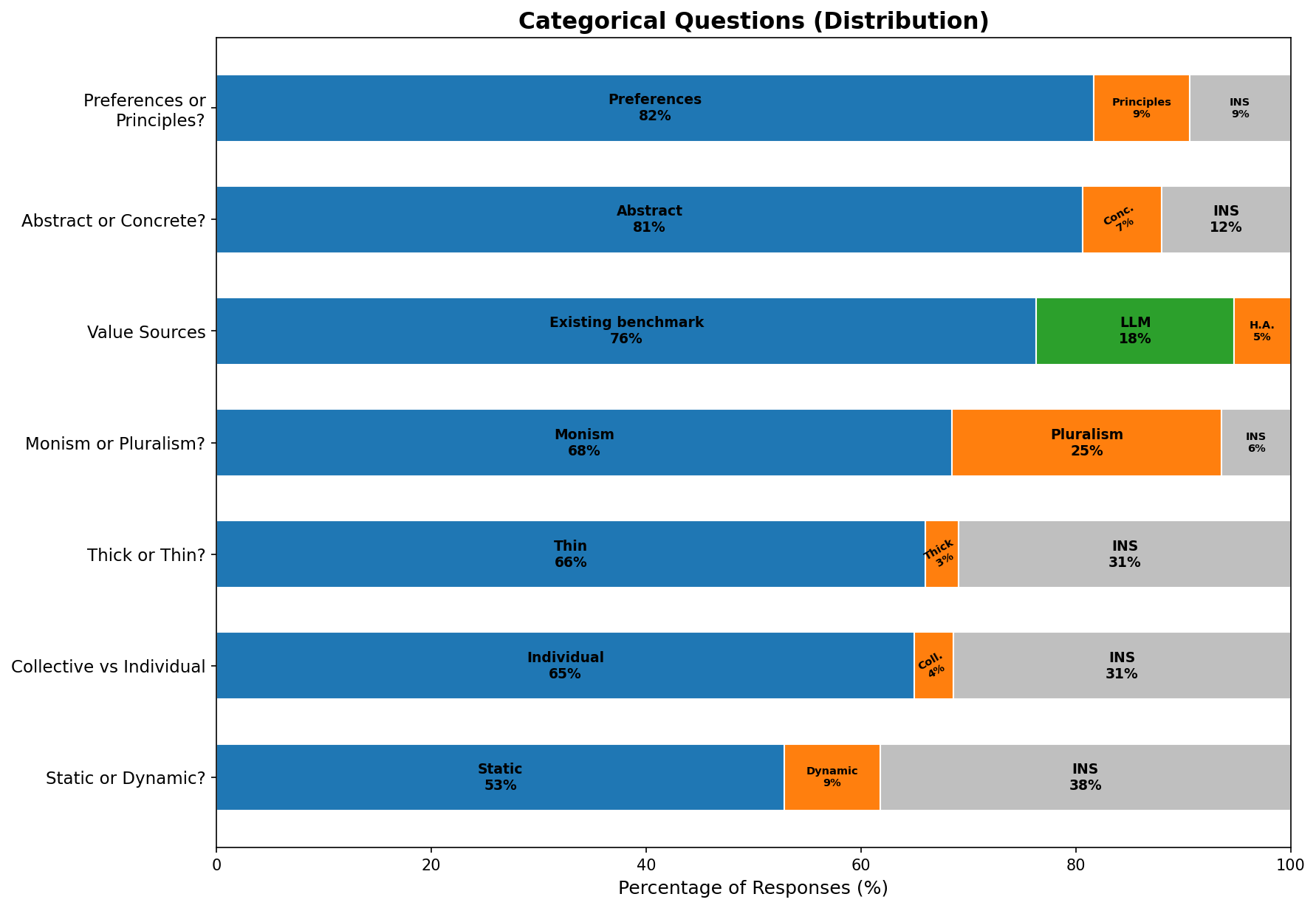}
    \caption{Summary of answers to our rubric aggregated across all raters for categorical questions. INS = "information not stated"}
    \label{fig:categorical_answers}
\end{figure*}

We present visualizations of the quantitative results in Figure 
\ref{fig:binary_answers} and Figure~\ref{fig:categorical_answers}. 

\subsection{Qualitative Analysis}
\label{qual_section}
We found that \textbf{79\% of the papers in our dataset neither defined nor described what they mean by "values"} despite ostensibly seeking to align models with values. In many papers, it was common to encounter the word "values" used interchangeably with "preferences" and "intentions," and as we address below, elision between values and preferences has become common in this subfield. In one rare example, values were also interchangeably used with "ideologies"~\cite{kirk2023personalisation}. In another, researchers trained a model on a corpus they built using Chinese laws and morality texts~\cite{xu2023align}. 

Despite largely not defining what values mean, 90\% of the papers treat values as measurable. This most often originated with different means of determining human preferences. Typically, researchers created datasets where LLMs were fed a prompt, generated two responses, and used either human raters, other LLMs (sometimes referred to as "autoraters")~\cite{lu2024online, zheng2024balancing,chakraborty2024maxmin,richemond2024offline}, or a mix of both~\cite{yu2023constructive,liu2024reward,wang2024secrets} to evaluate which of the two options better adhered to researchers' choice of alignment criteria; in other cases, one response might be treated as 'chosen' while the other is 'rejected.' Some papers drew from preference datasets and survey data that researchers who were not the authors of these papers generated~ \cite{zheng2024balancing,chakraborty2024maxmin,lou2024SPO,durmus2023toward,zhao2023group,miranda2024hybrid}, while others identified preferences they sought to optimize for including "conciseness..being humorous, philosophical, sycophantic, helpful, concise, creative, formal, expert, pleasant, and uplifting"~\cite{zhong2024panacea}. 

Preferences have become such a dominant framework that \textbf{82\% of the papers treat preferences as a stand-in for “values.”} By contrast, 9\% treated principles as a basis of values, for instance drawing from social choice theory \cite{siththaranjan2023distributional, chakraborty2024maxmin}. In papers where researchers created preference datasets through reliance on human annotators to rate LLMs' paired responses to prompts, only 29 of the 94 papers sampled provided a number for how many people comprised the rater pools. The number of raters tended to be $<100$ people, with the exception of one paper that used a 13,000-person rater pool~\cite{kopf2023openassistant}. When human rater pools were used, 77\% of that subset of papers \textbf{did not state the size of the rater pool}.

A growing debate within alignment is the extent to which pluralism of values can be achieved. One paper acknowledged that "different humans have different values, as it is nearly impossible to train a new large language model from scratch for individual preference"~\cite{zhou2024weak}. Despite this, most papers took individual rather than collective approaches to values. Rather than selecting values that are already practiced across societies, states, and other collectives, they rely on individual user interactions with LLMs (or LLM-to-LLM interactions), and in one case on collectives of 4-5 people~\cite{bakker2022fine}. In addition, \textbf{53\% of papers treated values as static}, whether implicitly or through direct acknowledgment of doing so despite recognizing that values evolve~\cite{liu2023training}.  

We found 66\% of papers presented “thin” descriptions and examples of what they meant by values. By contrast, “thick” descriptions (which accounted for 3 percent) draw from relational, observed values that are difficult to cleave apart from the lived context of the people who enact these values~\cite{geertz_thick_2008}. 

Likewise, \textbf{91\% of the papers did not refer to specific groups of humans whose values should be followed}, underscoring the taken-for-granted idea of values as universal. One paper noted "We claim that the approach described is agnostic to the ethical paradigm, the user’s preferences, and the legal or social framework, provided we can supply enough feedback"~\cite{leike2018scalable}, while another professed "No need to solve human values. We assume we do not need to solve hard philosophical questions of human values and value aggregation before we can align a superhuman researcher model well enough that it avoids egregiously catastrophic outcomes"~\cite{burns2023weak}. Another reckoned with the difficulties of treating preference datasets as representative: "This procedure aligns the behavior of GPT-3 to the stated preferences of a specific group of people (mostly our labelers and researchers), rather than any broader notion of 'human values'"~\cite{ouyang2022training}.

Finally, we noticed a trend in which a few papers suggest that AI models can develop their own value systems ~\cite{ngo2022alignment} and "internal meta-objectives"~\cite{phelps2023of}. Some researchers have begun to refer to models' "self-alignment"~\cite{guo2024human,li2024self} described as when "a new paradigm emerges where LLMs can achieve value alignment by themselves... transforming an unaligned LLM into one adhering to societal norms, independently of external resources" \cite{pang2024self}.

\section{Discussion}
\label{discussion}

Our systematic review of 94 research papers on AI value alignment reveals how the field is erasing human inputs from an endeavor whose success rests on purportedly steering AI systems to respect human values. Values are core components of how societies organize themselves; any attempt at value alignment is inextricable from society~\cite{graeber_toward_2001, haslanger_situated_2023, anderson_value_1995, zhi-xuan_beyond_2024, birhane_values_2022}. Recent social philosophy argues that societies are complex systems – or clusters of interacting systems – that reproduce themselves: their hierarchies, culture, and structures. These systems are also dynamic and constantly evolving~\cite{haslanger_situated_2023}. Yet we observe how research in this field abstracts away questions of whose values are represented, or what qualities constitute pluralism and dynamism in values, instead relegating these fundamental, unavoidable concerns to be taken up by others (if at all). The risk of repeatedly delaying explorations of what values are and how they can be accountably represented is that certain practices of design, deployment, and evaluation will ossify. At worst, this may result in normalizing widespread use of AI systems that have been trained on a narrow conception of value and claiming that this is the best the field can offer.  

The pressure to neglect these essential debates is reflected in a subset of papers that treat alignment as a step toward achieving "artificial superintelligence"~\cite{kim2024road}, or as an essential component of averting catastrophic and existential risks from AI~\cite{shen2023large, ngo2022alignment, burns2023weak}. This typifies the turn toward what we call "alignment without humans," or treatment of humans as an unreliable source of information on their own preferences ~\cite{miranda2024hybrid,wang2024secrets}, as too expensive \cite{wang2024cream,zheng2024balancing,liu2024reward} and time-consuming ~\cite{dai2023safe} to use, therefore justifying the use of synthetic means of preference elicitation, feedback, and evaluations ~\cite{hong2024adaptive,cui2024ultra,tian2024what}. We observe studies that rely on no human input for pre-determining value alignment criteria~\cite{mei2024hidden,chen2024low}, often through the use of simulated humans~\cite{chakraborty2023rebel,poddar_personalizing_2024,liu2023training}. Perhaps in part due to the desire to race toward solving "the alignment problem," one trend that arose from the corpus of papers is the dominance of Anthropic's Helpful, Harmless, Honest (HHH) approach. This ranged from papers that explicitly used the HH-RLHF dataset for training and/or evaluations ~\cite{yu2023constructive,lou2024SPO,huang2023catastrophic,zheng2024balancing,zeng2024token}, to others that name HHH as a guiding principle~\cite{ding2024ETA, gao2024HonestLLM, lu2024online, shen2023large,shi2024assessment}. 

Graeber's critique of applying economic theory to diverse cultures is apt for AI value alignment: "It [economics] also has the advantage of joining an extremely simple model of human nature with extremely complicated mathematical formulae that non-specialists can rarely understand, much less criticize."~\cite{graeber_toward_2001}. Our review of the literature reveals how RLHF and other technical alignment methods likewise join extremely simplistic models of human nature with complicated mathematical formulae that are difficult for non-specialists to understand or critique. 

Even though the majority of papers in our dataset do not define what they mean by values, adopting the framework of economic preference optimization to align models necessarily requires a commitment to a set of beliefs about what human values are. At a minimum this entails the belief that values are individual, rational, monist, static, maximizing utility functions, and abstract. The idea is that an agent that obeys some minimal conditions of rationality can be modeled as if the agent has an ordered set of preferences along with some probabilistic beliefs about what states of the world and itself will maximize the agent's utility (however defined) ~\cite{shea2018representation}. This particular set of assumptions about human nature and human values, drawn largely from economic decision theory and rational choice theory, has been empirically and theoretically challenged from multiple perspectives within and outside of economics~\cite{chibnik_anthropology_2011, graeber_debt_2011, graeber_toward_2001, gigerenzer1997bounded, frank_blind_2024} and AI alignment~\cite{bakker2022fine, kasirzadeh_plurality_2024, zhi-xuan_beyond_2024}. Rational or social choice theory do not describe the values or behavior of real human beings or cultures, but only the behavior of highly abstract, reductionist, and idealized fictional agents~\cite{frank_blind_2024}. 

Rather than asking "what do human beings actually care about?", rational choice theory asks - and presumes there is a normatively correct answer to - "what should human beings do in order to make the utility-maximizing decision?" This meta-theoretical commitment risks foreclosing alternative conceptions of human values, and limits the ability of AI systems to align to diverse cultural values~\cite{prabhakaran_human_2022, khan2025randomness, gabriel_artificial_2020}. Poddar \textit{et al.} point out, "Current RLHF approaches rely on a prescriptive set of values curated by a small set of AI researchers. Moreover, they typically assume that all end-users share the same set of values. Given the concerning lack of diversity in AI, it is clear that this approach cannot account for the range of social, moral, and political values that inform preferences in human populations"~\cite{poddar_personalizing_2024}. We call on the field to take seriously the fundamental weaknesses of utility maximization, rational choice theory and reductionist economic frameworks in AI alignment. 

\subsubsection{Utility maximization}

Economists and philosophers have generally assumed that people are trying to maximize \textit{something}: money, or love, or sometimes something else (most often, expected utility) with their choices \cite{glimcher2022efficiently, graeber_toward_2001}.Utility maximization treats all human behavior as the maximization of expected utility from a stable set of preferences, and holds that humans accumulate an optimal amount of information in \textit{markets}~\cite{chibnik_anthropology_2011, russell1995artificial, becker1976economic}. Thus, preferences are seen as being derived from a single utility function that each individual is somehow computing in their brain~\cite{gigerenzer1997bounded}. Taken to its extreme, utility theory holds that our behavior is entirely determined by maximizing this utility function, and preferences are assumed to not change substantially over time, nor to be very different between people of different socioeconomic classes, societies, and cultures~\cite{becker1976economic}. Finally, AI alignment assumes that an individual's reward function (and therefore their values), or preferences about the future, can be inferred, or approximated, by observing how humans behave by collecting data on how humans interact with AI outputs \cite{hadfield-menell_cooperative_2016}. 

Despite the empirical inadequacy of this ambiguous economic approach to AI and value alignment, creating an artificial agent that embodies economic theory has been the almost unquestioned north star in the field until recently~\cite{zhi-xuan_beyond_2024,gabriel_artificial_2020, russell_human_2019}. Influential work on value alignment explicitly argues that alignment should be formulated as a cooperative and interactive \textit{reward maximization process}~\cite{hadfield-menell_cooperative_2016}, with most alignment work implicitly adopting this approach~\cite{christiano2017deep, ouyang2022training, wang2024llms} the vast majority of technical approaches to alignment from our sample of papers implementing some form of utility maximization. These often unstated philosophical commitments to an economic worldview determine the methodology that alignment researchers use to measure and model human values.  We make this often implicit metatheorectical commitment explicit so that it can be critiqued and improved \cite{maxwell_comprehensibility_2005}.

We also evaluated our sample of papers for whether they adopted a framework of \textbf{utility maximization}, discovering that 87 \% of papers implicitly or explicitly use utility maximization as part of the technical framework to align models. Adopting the framework of utility maximization necessarily involves normative assumptions that are value-laden~\cite{graeber_toward_2001}. However, given that utility maximization is an inherent part of reinforcement learning from human feedback (RLHF) and reward modeling it is not surprising that the majority of alignment papers in our sample adopt this framework. This also stems from framing AI alignment in terms of the "principal-agent" framework from economic theory~\cite{phelps2023models, stanczak2025societal}. 

Moreover, utility maximization is cited as a property of models, e.g.,~\cite{mazeika2025utility} notes "it was shown that recent LLMs have structurally coherent, broad value systems. As they become more capable, their value systems increasingly conform to the axioms of utility theory, meaning they can be described as maximizing a utility function." Given that models are trained and fine-tuned according to the axioms of rational choice theory, should it be surprising that they behave according to utility theory? 



A small number of the papers recognize the inherent limitations of utility maximization. \cite{phelps2023models} points out: "Rather than seeking to impose a monolithic utility function on artificial agents, we propose a strategy of reducing information asymmetry and aligning interests through external incentives, much like the approach used in traditional economic solutions to principal-agent problems. "Furthermore, ~\cite{poddar_personalizing_2024} writes, "These insights suggest that human preferences are not derived from a single utility function, but are affected by unobserved, hidden user context." 

One of the starting points of our meta-review is \citet{zhi-xuan_beyond_2024}'s critique of the reliance on eliciting preferences as representations of human values, in which they ask: "What would AI alignment look like if it took these challenges seriously? It would move away from naive rational choice models of human decision making, towards richer models that include how we evaluate, commensurate, and act upon our values in boundedly rational ways. It would no longer take for granted expected utility theory, and instead explore systems for reasoning about the normativity of our preferences and values." 

\subsection{Alternative conceptions of value}
\label{alt_concepts}
\begin{quote}
\textit{"It's values all the way down"} 
- Alondra Nelson~\cite{nelson2023facct}
\end{quote}

The narrow economic conception of human values that determines the methodology and philosophy of alignment is by no means the only way to understand what humans value. As anthropologists point out, for 99 \% of humanity's history we did not live in market economies ~\cite{graeber_debt_2011}. Economic theory implicitly posits that we've always been capitalists and that capitalism is somehow natural ~\cite{hornborg1998ecological}. We argue that even recent work on pluralistic alignment is still fundamentally rooted in the orthodox economic framework. 

However, this is a distorted view of both societies and individual humans. Part of the challenge of rethinking alignment's approach to value comes from the social positioning of the field that 'doing anything is better than nothing,' without recognizing that over-utilizing narrow disciplinary approaches has the potential to make matters worse~\cite{dai2025artificiality}. A fundamental limitation of existing alignment methods is that they reinforce shallow behavioral dispositions rather than endowing LLMs with a genuine capacity for normative deliberation~\cite{milliere2025normative}.
We find that the abstractions and idealizations which alignment postulates to describe human values become confused for real phenomena.

Papers in our sample did in some cases engage directly with alternatives to utility theory, for example Huang et.al., (2025) who state, "We are interested in values not as abstract entities, but as operational priorities that influence how the system navigates its possible space of outputs." We expand others' calls to expand alignment approaches~\cite{casper2023open, lindstrom2024ai, zhi-xuan_beyond_2024} with our exploration of psychological and anthropological theories of value. 


\subsubsection{Cognitive or psychological theories of value}
Papers in our dataset referenced psychological theories of value that are accepted in mainstream psychology literature, such as Schwartz's theory of basic values \cite{sierra_value_2021, schwartz_overview_2012}. Schwartz theorizes value as a set of beliefs, closely linked to emotions, which inform the selection of actions. He further claims that there is a fixed set of "basic values" or value categories, which are \textbf{universal} across cultures, and tend to remain fixed over time within an adult individual. These values are conceptualized as guiding principles which can come into conflict, but are ultimately resolved through a hierarchy of importance. This leads to our characterization of the psychological theory of value as ultimately monistic. This individualistic view of values is challenged by empirical psychological and ethnographic work which shows a much more complex, situated, holistic and dynamic view of values \cite{chibnik_anthropology_2011, anderson_value_1995, haslanger_situated_2023}. This field is also complex, and we encourage engagement with existing debates in the cross-cultural study of values \cite{schwartz_overview_2012}. 

\subsubsection{Anthropological, situated theories of value}

The field with perhaps the most theoretical work and empirical data on human values - anthropology - has to date been ignored in AI alignment (cf.\ \cite{schmer2026culture}. Graeber suggests that what people call “value” may be thought of as “the way people represent the importance of their own actions to themselves, as reflected in one or another socially recognized form”~\cite{graeber_toward_2001}.  Economic theory's view of human nature is cynical, assuming that humans are entirely self-interested and will calculate the most effective way to obtain what they desire~\cite{becker1976economic}. These are values derived from living in market-based societies ~\cite{hornborg_technology_2014}. However, markets are not natural phenomena that arise spontaneously out of large complex societies as the economic narrative goes, but rather they require the enforcement of state violence to impose property rights, coercion, and the rule of law~\cite{graeber_debt_2011, mau2023mute}.

Economic anthropology recognizes that economic notions of value, articulated in objective monetary representations and utility theory, and ethical notions of value, articulated in moral terms, are “inextricably connected"~\cite{lambek2013value}. Graeber ~\cite{graeber_toward_2001} outlines three ways in which values have been theorized in the social sciences that go beyond the narrow conception of value in the economic sense: 
\begin{enumerate}

    \item "values" in the sociological sense: conceptions of what is ultimately good, proper, or desirable in human life
    \item "value" in the economic sense: the degree to which objects are desired, particularly, as measured by how much others are willing to give up to get them
    \item "value" in the linguistic sense, or how meaning is derived from the differences in the usage of words.
\end{enumerate}

However, in our dataset it is unusual for researchers to be specific about which humans, which values, and which cultures they target for alignment. We follow ~\cite{arzberger2024nothing}, who argue that values acquire situated meaning; hence, their concrete interpretations, means of actualization, and hierarchies vary between people depending on the situations in which they guide the notion of what people really value. This is closely aligned with situated-constructivist views of emotions, which see emotions not as basic universal categories, but as arising in specific socially- and culturally- salient circumstances. In other words, emotions are a category populated with highly variable instances~\cite{barrett2017theory}. Thus we can analogously we think of human values as a category populated with highly variable instances tied to specific cultural circumstances. 

\section{Limitations}
\label{limits}
One limitation of this work is the relatively small sample size of papers in our dataset compared with the vast number of research papers on this topic. This is especially challenging given the widespread practice of rapidly uploading research to arXiv, avoiding peer-review~\cite{gyevnar2025ai}. However, we attempt to correct for this by estimating the relative impact of our dataset on the field using both quantitative and qualitative measures. While we aimed to gather a representative subsample of available papers on AI value alignment, we make no claims to the completeness or comprehensiveness of the sampled papers. The sample size was also limited due to time constraints on our small team of co-authors. We also acknowledge that the scoring rubric has many limitations. The questions are highly nuanced and difficult to boil down to a binary or even ternary answer. We considered a multi-point Likert scale, which was ultimately abandoned due to concerns about calibrating relative scales.


\subsection{Implications for future work}

We suggest building on our analysis to apply concepts from humanistic social sciences such as anthropology and psychology to operationalize human values in AI in a more representative and holistic way that explicitly accounts for: concreteness/abstraction, the thick/thin distinction, distinctions between monism and pluralism, and embracing a view of values not as abstract universal things human possess, but as constructed in specific social and cultural contexts~\cite{kasirzadeh_plurality_2024, sorensen_roadmap_2024, benkler2023assessing, arzberger2024nothing}.

To deepen the critique of "preferentism," we also recommend an examination of the content of prompts and paired responses in preference training datasets, as many contain largely nonsensical conversations with an LLM and involve choosing the less absurd of two unrealistic examples in each pair. Communicating the shaky foundations of these unquestioned defaults is of particular value to informing policymakers and the public about the limits of value alignment. Moreover, it can contribute to more effective sociotechnical and policy measures derived from realistic conceptions of AI's risks.



\section{Conclusion}
We began with an examination of the theory of value(s) to which AI value alignment research subscribes, drawing from interdisciplinary philosophy and social scientific research on human values to develop a schema for answering this question. We argued that AI alignment lacks a concrete theory of human values and implicitly adopts beliefs about human values rooted in economic theory. Through our review, we have shown that the assumptions guiding technical AI alignment practices limit the possible range of human values with which AI can be aligned by translating alignment into utility theory. As Lisa Feldman Barrett argues, "Even a scientist who believes they are purely driven by data is doing philosophy"~\cite{barrett2025s}. Tech companies and AI value alignment researchers enact philosophy every day as they engage in value alignment work when choosing training data, setting up model architectures, and especially when running RLHF experiments. Having laid bare AI value alignment’s philosophical commitments, we invite researchers from a wider variety of disciplines to bring greater specificity and under-explored perspectives into the debate on whether and how AI can address human values. 

\section{Generative AI Usage Statement}
Generative AI was not used in the research or writing of this publication. 






\bibliography{alignmentwohumans}


\end{document}